\documentclass[manuscript]{acmart}

\AtBeginDocument{%
  }

\setcopyright{acmlicensed}
\copyrightyear{2025}
\acmYear{2025} 
\acmDOI{xxxx.xxxxx}

\usepackage{amssymb}
\usepackage{amsmath}
\usepackage{graphicx} % Required for inserting images
\usepackage{forest}
\usepackage{xcolor}
\usepackage{svg} 
\usepackage{amsmath}
\usepackage{amssymb}
\usepackage{pgf-pie}
\usepackage{subcaption}
\usepackage{float} 
\usepackage{tikz}
\usepackage{arydshln}
\usepackage{multicol}
\usepackage{multirow}
\usepackage{adjustbox}
\usepackage[utf8]{inputenc}
\usepackage{tabularx}
\usepackage{array} % Include in the preamble
\usepackage{booktabs}
\usepackage{hyperref}
\usepackage{natbib}

\usepackage{pifont}% http://ctan.org/pkg/pifont
\newcommand{\cmark}{\ding{51}}%
\newcommand{\xmark}{\ding{55}}%

\newcolumntype{R}[1]{>{\raggedleft\arraybackslash }b{#1}}
\newcolumntype{L}[1]{>{\raggedright\arraybackslash }b{#1}}
\newcolumntype{C}[1]{>{\centering\arraybackslash   }b{#1}}
\newcommand{\NbSurveyed}{$ 45\ $} 

\begin{document}

%%
%% The "title" command has an optional parameter,
%% allowing the author to define a "short title" to be used in page headers.
\title{Arabic Multimodal Machine Learning: Datasets, Applications, Approaches, and Challenges
}\thanks{ This work is supported by the Directorate-General for Scientific Research and Technological Development and performed under the PRFU Projects:C00L07N030120220002 and C00L07UN470120230001.
}
%%
%% The "author" command and its associated commands are used to define
%% the authors and their affiliations.
%% Of note is the shared affiliation of the first two authors, and the
%% "authornote" and "authornotemark" commands
%% used to denote shared contribution to the research.

\author{Abdelhamid Haouhat}
%\email{a.haouhat@lagh-univ.dz}
\orcid{0009-0008-0820-5381}
\affiliation{%
  \institution{Université Amar Telidji}
  \city{Laghouat}
  \state{Laghouat}
  \country{Algeria}
}
\author{Slimane Bellaouar}
\orcid{0000-0001-8357-5501}
%\email{bellaouar.slimane@univ-ghardaia.dz}
\affiliation{%
  \institution{Université de Ghardaia}
  \city{Ghardaia}
  \country{Algeria}}

\author{Attia Nehar}
\orcid{0000-0002-3245-6913}
\affiliation{%
  \institution{Ziane Achour University}
  \city{Djelfa}
  \country{Algeria}
}
\author{Hadda Cherroun}
\orcid{0000-0002-5117-0320}
%\email{hadda\_cherroun@lagh-univ.dz}
\affiliation{%
  \institution{Université Amar Telidji}
  \city{Laghouat}
  \state{Laghouat}
  \country{Algeria}
}
\author{Ahmed Abdelali}
\orcid{0000-0002-4160-8181}
%\email{ }
\affiliation{%
  \institution{Humain}
  \city{Ryiad}
  \country{KSA}
}

%% ,screen,review
%% By default, the full list of authors will be used in the page
%% headers. Often, this list is too long, and will overlap
%% other information printed in the page headers. This command allows
%% the author to define a more concise list
%% of authors' names for this purpose.
\renewcommand{\shortauthors}{}

%%
%% The abstract is a short summary of the work to be presented in the
%% article.
\begin{abstract}
%% Text of abstract
Multimodal Machine Learning (MML) aims to integrate and analyze information from diverse modalities, such as text, audio, and visuals, enabling machines to address complex tasks like sentiment analysis, emotion recognition, and multimedia retrieval. Recently, Arabic MML has reached a certain level of maturity in its foundational development, making it time to conduct a comprehensive survey. This paper explores Arabic MML by categorizing efforts through a novel taxonomy and analyzing existing research. Our taxonomy organizes these efforts into four key topics: datasets, applications, approaches, and challenges. By providing a structured overview, this survey offers insights into the current state of Arabic MML, highlighting areas that have not been investigated and critical research gaps. Researchers will be empowered to build upon the identified opportunities and address challenges to advance the field. %An associated GitHub repository collecting the latest
%papers and repos is available at \href{https://github.com/...}{https://github.com/******/}
\end{abstract}
%%
%% The code below is generated by the tool at http://dl.acm.org/ccs.cfm.
%% Please copy and paste the code instead of the example below.

%%
%% Keywords. The author(s) should pick words that accurately describe
%% the work being presented. Separate the keywords with commas.

%%
%% This command processes the author and affiliation and title
%% information and builds the first part of the formatted document.
\maketitle

\section{Introduction}
Humans naturally perceive and interpret the world through a combination of senses such as sight, hearing, and touch. Without the integration of these multiple senses, understanding the environment can become challenging. For example, sarcasm cannot be fully identified through spoken words alone; it requires additional context, such as speech tone and facial expressions. This demonstrates the essential role of multiple modalities in human understanding and expression. 

Inspired by this natural multisensory perception, Multimodal Machine Learning (MML) integrates information from diverse modalities to address complex tasks such as sentiment analysis, emotion recognition, and multimedia retrieval. These modalities are represented in various ways~\cite{10.5555/3104482.3104569}. For the textual modality, word or sentence embeddings are commonly used to capture the semantic meaning of the text. In contrast, for visual and acoustic modalities, researchers employ statistical methods or learned models to represent intricate and nuanced features, such as facial expressions, gestures, and speech tone that convey essential information~\cite{10.1109/TPAMI.2018.2798607}. Furthermore, to effectively leverage the common information across these diverse modalities, researchers have introduced techniques such as fusion methods and alignment strategies~\cite{10.1145/3656580,10041115,10.1109/TPAMI.2018.2798607}. These approaches aim to integrate and align multimodal data, enabling models to learn richer and more cohesive representations.

The gap in Arabic MML research stems from unique challenges in the Arabic context, such as its rich linguistic diversity, morphological complexity, and the scarcity of multimodal annotated datasets. At the same time, the Arabic language presents unique opportunities for innovation, particularly in applications like sentiment analysis, question answering, and more, where linguistic nuances play a critical role. Addressing these challenges and leveraging these opportunities are crucial to advance Arabic MML.

Early studies in multimodal processing relied on traditional methods that employed heuristic processes~\cite{7944543}. These methods used predefined rules crafted by experts to process and integrate multimodal data. While these methods were effective for basic tasks, they exhibited significant limitations, including reliance on manual feature engineering, dependency on human interaction, and a cumbersome implementation process. Such challenges highlighted the need for more automated and scalable solutions.

Subsequently, classical machine learning approaches were introduced to address the shortcomings of traditional methods. These methods, exploit Support Vector Machines (SVMs)~\cite{9148603}, Hidden Markov Models (HMMs), K-Nearest Neighbors (KNN)~\cite{s_alaa2015}, and Multilayer Perceptrons (MLPs)~\cite{10.1007/s10772-022-09981-w,bhatia-etal-2024-qalam}, leveraged statistical and mathematical techniques to process multimodal data. By enabling automated feature selection and learning from structured datasets, classical machine learning methods achieved competitive results in tasks such as image retrieval and speech recognition. However, their limited scalability in handling sequential data restricted their effectiveness in more complex multimodal scenarios.

The advent of deep learning revolutionized multimodal machine learning by enabling more effective handling of diverse data types and complex interactions between modalities. For sequential data, such as audio signals or visual frames, researchers adopted Recurrent Neural Networks (RNNs) and their variants~\cite{Elbedwehy2023,cmc.2024.048104}, which excel at capturing temporal dependencies. Convolutional Neural Networks (CNNs)~\cite{Zater2022BenchMarkingAI,ALROKEN2023103005,8717419} proved effective for processing spatial features in images and videos. More recently, Transformers~\cite{10.5555/3295222.3295349} have emerged as state-of-the-art models for sequential data, leveraging attention mechanisms to capture long-term dependencies. Many studies have incorporated attention-based and cross-attention modules to enhance the ability of models to learn nuanced correlations between modalities~\cite{zaytoon-etal-2024-alexunlp,ALHARBI2024102084, Mehra2024-gc}.
These advancements in Transformers have paved the way for cutting-edge Large Language Models (LLMs), which have been revolutionized by the multimodal research perspective and showing remarkable advancements in tasks such as question answering, image generation, and image descriptions~\cite{bhatia-etal-2024-qalam,alwajih-etal-2024-dallah,alwajih-etal-2024-peacock}. Additionally, LLMs have shown exceptional performance in zero-shot and few-shot learning scenarios, generalizing to new tasks with minimal additional training. Simultaneously, researchers are addressing challenges in low-resource settings and continual learning by developing methods to reduce the memory and computational demands of these models, ensuring their scalability and accessibility~\cite{marouf2025imagesproblemretainingknowledge, Mehra2024-gc}.

This survey bridges the gap in Arabic MML research by offering a comprehensive overview of the field. We introduce a novel taxonomy that categorizes existing efforts along four dimensions: data-centric aspects, application-specific domains, methodological approaches, and key challenges. Additionally, we analyze current obstacles and provide insights into these critical areas of Arabic MML. Through this work, our aim is to bridge the gap between the growing demand for MML applications and the current state of Arabic MML research.

The remainder of the paper is organized as follows. 
In Section~\ref{sec:Taxonomy}, we present the methodology and introduce our proposed taxonomy for Arabic MML efforts. Section~\ref{sec:Review} provides a comprehensive review of existing MML studies. Section~\ref{sec:Discussion} offers an in-depth discussion of the findings and their implications. Finally, Section~\ref{sec:conclusion} concludes the paper with insights and directions for future research.

\section{Methodology and Taxonomy }
\label{sec:Taxonomy}

 We have reviewed \NbSurveyed papers and studies on various aspects of Arabic Multimodal Machine Learning research. We have developed a taxonomy based on several classification criteria to present these studies effectively. The primary focus is to identify the key patterns and trends within the field.     

 The first criterion focuses on dataset-centric contributions, which play a key role in advancing research in this area. These studies are described based on their intended applications (e.g., sentiment analysis, emotion recognition) and the types of modalities they involve (e.g., text, speech, image, video).

Furthermore, we classify Arabic MML efforts based on their target applications, including but not limited to Sentiment Analysis, Emotion Recognition, Propaganda Detection, Speech Recognition, Remote Sensing, Image and Video Retrieval, Video Classification, and other specific applications such as multimodal interaction systems and accessibility tools.

Additionally, we analyze the techniques employed in Arabic multimodal research, spanning traditional methods, classical machine learning,  and advanced deep learning architectures, including CNNs, RNNs, Transformers, and multimodal-specific models. 

Finally, under the last classification criterion, we explore the challenges these studies addressed, with a particular focus on modality representation. This includes examining how Arabic multimodal data are represented and processed, as well as the techniques used to extract meaningful features from modalities such as text, speech, images, and video.
We also explore fusion techniques, emphasizing methods for integrating information across multiple modalities, including early fusion, late fusion, cross-attention mechanisms, and hybrid approaches. In addition, we discuss the translation of input modalities into other output modalities and the alignment between relevant input modalities to ensure seamless integration.

This taxonomy is the foundation for our detailed review, offering a systematic framework to describe, analyze, and compare Arabic multimodal research efforts. It provides a comprehensive understanding of the advancements, limitations, and potential future directions in Arabic multimodal research.

\begin{figure}[h!]
    \centering
       \includegraphics[width=\textwidth]{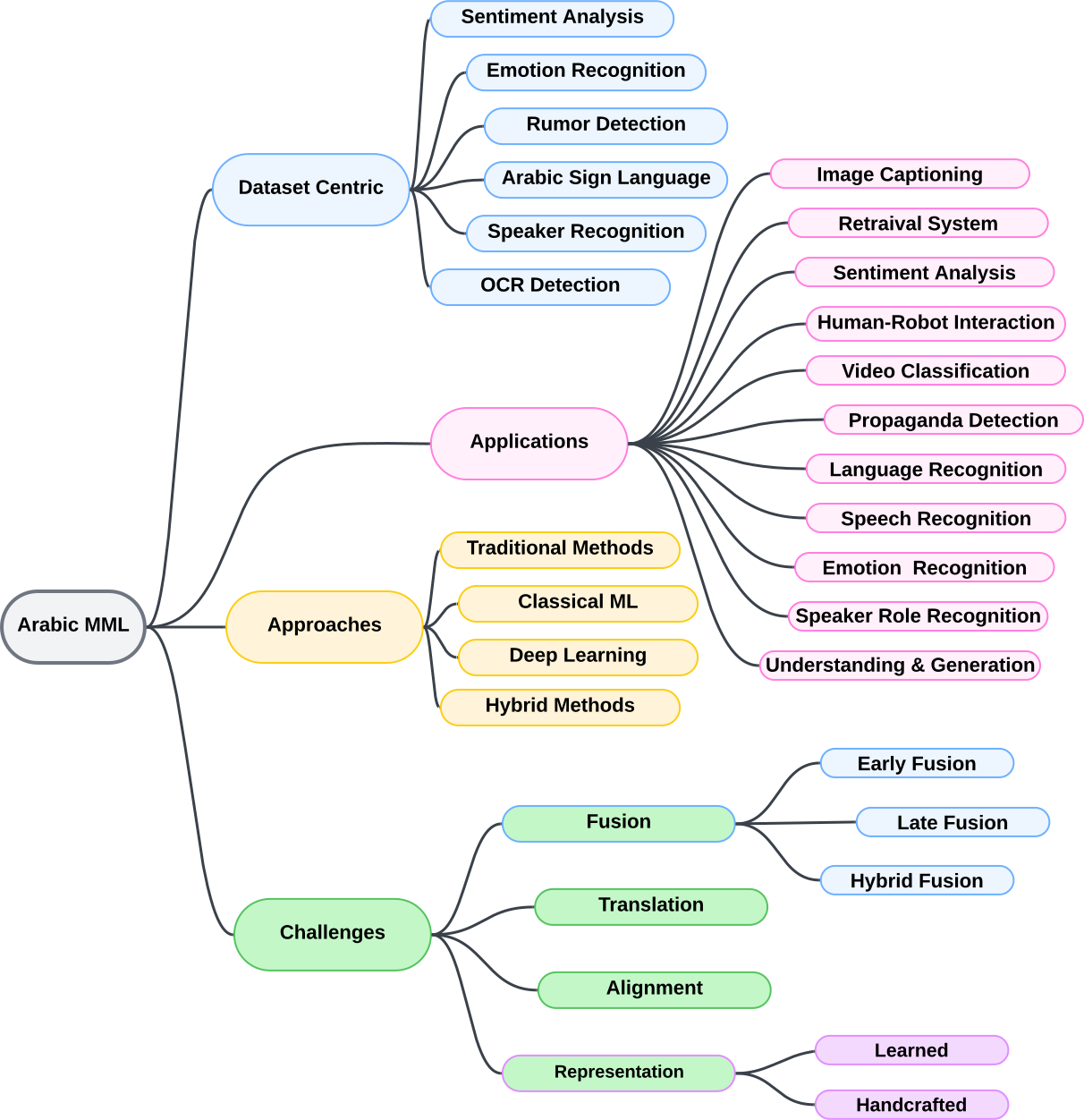}% Replace 'taxonomy_diagram' with your SVG filename
    \caption{An Overview of the Proposed Taxonomy for Arabic Multimodal Machine Learning Research.}
    \label{fig:taxonomy}
\end{figure}

Figure~\ref{fig:taxonomy} provides a visual representation of our proposed taxonomy, illustrating the categorized research efforts in Arabic multimodal studies across target applications, core challenges, machine learning approaches, and dataset-centric contributions.

\newpage
%%%%%%%%%%%%%%%%%%%%%%%%%%%%%%%%%%%%%%%%%
\section{Review of MML Work}
\label{sec:Review}
In this section, we give a full study of our taxonomy, structured around four key topics: data-centric aspects, applications, approaches, and challenges. Each of these key topics will be examined in detail, offering a holistic grasp of the taxonomy and its ramifications. By delving into these dimensions, we seek to present a clear and methodical framework that not only organizes existing information but also offers potential for future study and innovation.
%%%
\subsection{Dataset Centric}

Data is a fundamental pillar of the machine learning era, serving as the foundation for learning.  This importance has led researchers to focus on the creation of datasets for diverse machine learning tasks in the Arabic context. 
One of the earliest efforts in building multimodal Arabic datasets was undertaken by Ashraf et al.~\cite{AbdelRaouf2010-qo} in $2010$ with the introduction of their Multimodal Arabic Corpus (MMAC). This later dataset was specifically designed to support Optical Character Recognition (OCR) development and linguistic research. The corpus includes a variety of Arabic text images, encompassing real-world, computer-generated, and noisy examples. To enhance the dataset's flexibility and richness, the authors incorporated not only full word sequences but also partial word sequences, categorized as 'Piece of Words' and 'Naked Piece of Words.'

Subsequently, Samar et al.~\cite{Samar2013} created the Audio-Visual Speech Recognition (AVAS) corpus to address audio-visual recognition tasks.  Introduced in 2013, this dataset incorporates variations in illumination and head pose, making it suitable for advancing multimodal recognition systems. The AVAS corpus is designed to support the development of audio-visual speech recognition (AVSR) systems, audio-only speech recognition, lip-reading (visual-only) systems, and face recognition under diverse conditions. Its variability in illumination and pose ensures the dataset's utility for enhancing the performance and robustness of speech recognition systems in dynamic, real-world environments. 
 
Sentiment Analysis and Emotion Recognition are among the most popular and captivating topics in research. In $2019$, Alqarafi et al.~\cite{Alqarafi2019} introduced the Arabic Multi-Modal Dataset (AMMD) for sentiment analysis, comprising $830$ segments, each containing visual and language modalities along with sentiment labels. The dataset is labeled with four classes: objective, subjective, positive, and negative. The authors utilized Support Vector Machine (SVM)-based classification models to perform sentiment analysis on this dataset. Expanding on sentiment analysis datasets, Haouhat et al.~\cite{Haouhat-IDSTA-23} developed the Arabic Multimodal Sentiment Analysis (AMSA) dataset, a larger and more comprehensive resource. AMSA integrates all three available modalities commonly found in multimedia videos: visual, auditory, and textual transcriptions. It includes $60$ long videos and $540$ segments, providing a rich dataset for multimodal sentiment analysis.

Focusing on emotion recognition, AbuShaqra et al.~\cite{Shaqra2019} conducted a study in $2019$ and introduced the AVANEmo dataset for Audio-Visual Natural Emotions. This dataset comprises $3000$ clips containing both audio and video data, annotated with six basic emotional labels: Happy, Sad, Angry, Surprise, Disgust, and Neutral. The authors also evaluated the performance of their emotion recognition models, achieving notable accuracy using both audio and visual modalities.

In 2021, Bellagha et Zrigui~\cite{BELLAGHA202159} developed a multimodal dataset specifically designed for recognizing speaker roles in Arabic TV broadcasts. The dataset, created using data from the Multi-genre Broadcast  MGB-2 Challenge, comprises approximately  $205$ hours of audio data with corresponding transcriptions, and $8,112$ pairs of annotated speaker and speech turns. It is annotated for both text and audio modalities to identify speaker roles during TV show interactions. The authors evaluated the dataset using baseline classifiers for both audio and text, highlighting the significance of integrating both modalities to accurately identify speaker roles in television shows. 
The dataset is further augmented by the AraFACT dataset~\cite{sheikh-ali-etal-2021-arafacts} containing $1,726$ samples.

In 2022, Albalawi et al.~\cite{Albalawi2022} introduced a carefully curated dataset of $2,299$ samples, along with a novel approach to rumor detection. The dataset combines two modalities -text and images- extracted from social media posts, providing a valuable resource for advancing research in this domain.

In 2023, Luqman Hamza~\cite{10.1109/FG57933.2023.10042720} introduced a novel Dataset and Benchmark, ArabSign, for Arabic Sign Language (ArSL). ArabSign contains 9,335 samples recorded from six signers, amounting to approximately 10 hours of recorded sentences. Each sentence has an average length of 3.1 signs. The recordings were captured using a Kinect V2 camera, providing three types of information--color, depth, and skeleton joint points-- all recorded simultaneously for each sentence.

Samah Abbas et al.~\cite{ABBAS2024102165} later developed a multimodal dataset for ArSL, specifically in religious speech, such as Friday Sermons. The dataset comprises $1,950$ audio signals, 131 corresponding text phrases, and $262$ annotated ArSL videos. Annotations made using the ELAN tool based on gloss representation.  Focusing on formal speech, this dataset provides a valuable resource for multimodal learning models that can bridge the gap between the audio, text, and sign language modalities. To evaluate annotation consistency, the authors employed the Jaccard Similarity metric between gloss representations across two different signers, achieving a similarity score of 85\%.

CAMEL-Bench~\cite{ghaboura2024camelbenchcomprehensivearabiclmm} is a novel benchmark developed by researchers at MBZUAI\footnote{https://mbzuai.ac.ae/} to evaluate the Visual question-answering (VQA) capabilities of Arabic Language Multimodal Models (LMMs).  This benchmark comprises over 29,000 questions across eight diverse domains and 38 sub-domains, meticulously curated by native Arabic speakers to ensure cultural and linguistic relevance.  CAMEL-Bench assesses LMM performance on a variety of tasks, including OCR, medical imaging analysis, remote sensing interpretation, and understanding culturally specific content. 
%%%%
Table~\ref{tab:arabic_datasets} provides an overview of key multimodal Arabic datasets, highlighting their modalities and applications.
\begin{table}[H]
\centering
\caption{The table summarizes various Arabic multimodal datasets, detailing their mod(alities), app(lications), and avail(ability). The \textbf{mod} column uses the following abbreviations: A(Audio), T(Text), V(Video), I(Image), C(Class), D(Depth), S(Skeleton), and SP(Social Media Post). \textbf{app/features } describes the dataset’s use cases. \textbf{avail} column  indicates whether the dataset is publicly available(\textcolor{green}{\cmark}), unavailable (\textcolor{red}{\xmark}), or uncertain (N/A)}
\begin{tabular}{ L{3.5cm}  L{1.5 cm}  L{7cm} C{0.6cm}}
\toprule
\textbf{dataset}  & \textbf{mod} & \textbf{app/ features} &  \textbf{avail}\\ \midrule
MMAC~\cite{AbdelRaouf2010-qo} (2010)   & $T,I$& OCR development, linguistic research & \textcolor{red}{\xmark} \\ \cdashline{1-4}
 
AVAS~\cite{Samar2013} (2013)  & $A,V$ & based on AVSR, lip-reading, Face Recognition under varying conditions & \textcolor{red}{\xmark}
 \\ \cdashline{1-4}
 
AMMD~\cite{Alqarafi2019} (2019)  & $T,V$ & Sentiment Analysis & \textcolor{red}{\xmark} \\ \cdashline{1-4}
 
AVANEmo~\cite{Shaqra2019}  (2019) & $A,V$& Emotion Recognition (6 basic emotions) &  N/A \\ \cdashline{1-4}
 
Bellagha et Zrigui~\cite{BELLAGHA202159} (2021) & $A,T$ & Speaker Role Recognition in TV broadcasts & \textcolor{green}{\cmark} \\ \cdashline{1-4}
 
Albalawi et al.\cite{Albalawi2022} (2022) & $T,SP$ & Rumor Detection &   N/A  \\ \cdashline{1-4}
 
ArabSign~\cite{10.1109/FG57933.2023.10042720} (2023) & $A,V,D,S$ & Arabic Sign Language (ArSL) Recognition & \textcolor{green}{\cmark}  \\ \cdashline{1-4}
 
AMSA~\cite{Haouhat-IDSTA-23}  (2023) & $A,T,V$& Sentiment Analysis &  \textcolor{green}{\cmark}\\ \cdashline{1-4}
 
Abbas et al.\cite{ABBAS2024102165} (2024) & $A,T,V$ & Religious Speech, ArSL annotation &   N/A  \\ \cdashline{1-4} 

Alroken~\cite{ALROKEN2023103005} (2023)&  $A,V$& Emotion Recognition & \textcolor{red}{\xmark}  \\ \cdashline{1-4}
 L2-KSU~\cite{Alrashoudi2025}
 (2025)& $A,T$  & Speech Recognition, Mispronunciation Detection &  N/A  \\
\cdashline{1-4} 
CAMEL-Bench.\cite{ghaboura2024camelbenchcomprehensivearabiclmm}(2025)&  $T,I,C$ & 45 Visual QA  datasets used for benchmarking LLM, available at
\href{https://huggingface.co/collections/ahmedheakl/camel-bench-670750f3998395452cd3b7b1}{HuggingFace}  & \textcolor{green}{\cmark}
\\ \bottomrule
\end{tabular}

\label{tab:arabic_datasets}
\end{table}

\subsection{Applications} 
Arabic MML frameworks have explored a wide variety of applications, from the most popular, such as sentiment analysis and emotion recognition, to more complex ones, such as computer vision image captioning and human-robot interactions. Hereafter, we categorize the reviewed studies based on their respective application domains.

%#################################### 
\subsubsection{Sentiment Analysis}  Researchers have shown that analyzing a speaker's tone alongside his facial expressions enables models to understand better emotional states, such as distinguishing sarcasm from sincerity. For example, a smiling face paired with a seemingly angry tone might indicate sarcasm, a nuance that single-modality systems often fail to capture~\cite{10.1109/TPAMI.2018.2798607,shi2025comprehensivesurveycontemporaryarabic,9677274,PORIA201798, Haouhat-IDSTA-23}. This integration reflects the human ability to infer sentiments more accurately by combining visual and auditory cues.

In 2020, Al-Azani and El-Alfy~\cite{9148603} presented a novel approach to multimodal sentiment analysis, focusing specifically on Arabic content. They extract features from each modality: \textit{word embeddings }for textual data, \textit{prosodic} and \textit{spectral} features for auditory data, and face optic flow motion for visual data. Additionally, the authors developed their dataset for experimental evaluation.
A key contribution of this work is the introduction of a hybrid multi-level fusion approach, which combines feature-level, score-level, and decision-level fusions to integrate information across modalities effectively. For the final classification stage, the authors utilize Support Vector Machines (SVM) with a linear kernel (using LibSVM) and logistic regression (LR) with L2 norm regularization and the Liblinear solver to classify sentiments.
While the hybrid multi-level fusion approach represents a thorough attempt to leverage the strengths of various fusion types, it relies relatively on traditional feature extraction techniques and classifiers. Although effective for its time, these methods may not fully capitalize on recent advances in deep learning architectures, which have demonstrated superior performance in multimodal sentiment analysis. 
Youcef and Barigou~\cite{9677274} investigated both unimodal and multimodal sentiment analysis emphasizing the growing prevalence of multimodal sentiment expression through images, videos, and audio. Their study highlights the importance of exploring these modalities. 
In 2023, Alalem et al.~\cite{10453875} introduced ATFusion, a model that combines CNN, LSTM, and transformers as local classifiers for text and audio data, integrating modalities using the Group Gated Fusion (GGF) technique. Evaluated on the IEMOCAP and EYASE datasets, the model demonstrated promising results in sentiment analysis for both English and Egyptian Arabic, showcasing its effectiveness in multilingual and multimodal scenarios. While the GGF fusion and the focus on dialectal Arabic are innovative, increasing dataset diversity could further improve the model’s generalizability.

Recently, Shi and Agrawal~\cite{shi2025comprehensivesurveycontemporaryarabic} conducted a detailed survey of contemporary Arabic sentiment analysis systems, ranging from unimodal to multimodal approaches.

\subsubsection{Emotion Recognition} Many researchers regard sentiment analysis and emotion recognition as closely related fields, thus we can find some effort work on both tasks such as Amd'SaEr dataset~\footnote{https://github.com/belgats/AMSAER}~\cite{Haouhat2024AMDSAER}. Although Alroken and Balras~\cite{ALROKEN2023103005} proposed an emotion recognition-only Audio-Visual model that integrates  Mel-Frequency Cepstral Coefficients fed to CNN for audio feature extraction and VGG for visual feature extraction. They employed concatenation for fusion and utilized a softmax function for emotion prediction. The study explored both unimodal and multimodal training approaches, assessed the impact of dataset size on visual classifiers, and examined the effects of modality interaction and overlapping segmentation. 
Abu Shaqra et al.~\cite{10.1007/s10772-022-09981-w} employed audio and visual modalities with a decision fusion strategy, utilizing binary and 6-class MLP models for audio classification and LSTM models for visual classification. The final classification was achieved by combining the output probabilities from both modalities. The study also introduced a gender-based model, processing features separately for male and female subjects, leveraging OpenSMILE for audio and CNN for visual features. 
The use of the specialized AVANEmo dataset and the incorporation of gender-based modeling offer valuable insights into Arabic emotion recognition, though the dataset’s limited size and lack of accessibility hinder reproducibility and may limit the generalization of the findings.

\subsubsection{Image Captioning} Elbedwehy and Medhat~\cite{Elbedwehy2023} tackled the architectural design challenges of cross-modal tasks, specifically transitioning from image to text modality. Their proposed model employed a transformer-based architecture that extracts image features using three pre-trained models (XCIT, SWIN, and ConvNeXT), which are concatenated and encoded through fully connected layers to form the encoder unit. On the textual side, pre-trained language models (AraBERT, AraELECTRA, MARBERTv2, or CamelBERT) are utilized to generate word embeddings, which are then processed by an LSTM decoder to produce output tokens.
To evaluate their approach, the authors conducted experiments on two widely used datasets, Flickr8k and MSCOCO, comparing their method against existing models. Using BLEU-1 to BLEU-4 metrics as the primary evaluation criteria, their method demonstrated improved performance for Arabic image captioning tasks.
Recently, Emran and Dan~\cite{cmc.2024.048104}  introduced a novel system for generating cross-lingual image descriptions, focusing on producing semantically coherent and culturally aligned translations from English to Arabic. 
Their approach combines a neural network model with a semantic matching module to ensure both semantic and stylistic alignment. The system employs a multimodal semantic matching module, where input images are encoded using a pre-trained ResNet followed by fully connected layers (FCL). The resulting image features are then decoded into Arabic captions using an LSTM. To ensure cross-modal coherence, the model calculates semantic similarity between the mapped representations of the generated sentence and the image within a common knowledge space. This process is further enhanced by a language evaluation module that incorporates rewards based on cross-modal semantic matching and language evaluation. 
For evaluation, the authors introduced a dedicated dataset, AraImg2K, consisting of $2,000$ images, each annotated with five captions. While the proposed system does not introduce an efficient fusion method for integrating image-text modalities, it effectively addresses the challenges of modality alignment and the translation of image data into textual description.
Muhy Za'ter and Talafha~\cite{Zater2022BenchMarkingAI} explored the use of multi-task learning to improve Arabic image captioning by integrating various textual representations, such as AraVec, ELMo-based embeddings, and BERT-based models, with image features extracted using ResNet and VGGNet. 
These features are processed through CNN-based shared layers to learn common representations for multiple tasks. For object recognition and action classification, fully connected layers and a SoftMax predictor are employed to handle categorical outputs. LSTM layers are utilized to process and generate variable-length sequences effectively for the image captioning task. The study highlights that leveraging multi-task learning and pre-trained word embeddings significantly enhances image captioning performance by utilizing shared knowledge across tasks.

\subsubsection{Retrieval Systems} Retrieval systems are one of Multimodal Arabic research's most commonly addressed tasks. Karkar et al.~\cite{7944543} presented a multimedia system designed for educational purposes, particularly for children. The system allows users to input textual content and retrieve relevant illustrative images. It comprises five main components: an image repository, a knowledge base, text processing, conceptual graph visualization, and visual illustration. The image repository stores visual materials collected from the internet for efficient retrieval. The knowledge base organizes content ontologically, combining lexical and educational resources. Text processing involves extracting relationships between words through an entity-word matrix, which is then used to generate a conceptual graph linked to the text. Finally, the system maps existing visual materials to semantic conceptual graphs to retrieve illustrative images that match the textual input.
Salah and Aljaam.~\cite{8717419} introduced a modular system designed to support children with learning difficulties by transforming user-provided textual stories into corresponding descriptive visual elements, such as images and animations. The system extracts keywords from the story text, selects relevant images, and generates captions for those images using pre-trained CNN and LSTM models. It then evaluates the similarity between the input text and the generated image captions, ranging sentences based on similarity scores to identify the most relevant matches.
While this work effectively employs statistical methods to process textual inputs, there is a need for further automation of the workflow to reduce reliance on human interpretation, thereby enhancing scalability and efficiency. 
Moselhy et al.~\cite{10278299} developed a retrieval and recommendation system for Arabic content that utilizes multimodal data to enhance search accuracy and user experience. The system combines textual and visual modalities, focusing on aligning features extracted from both using advanced fusion techniques. These techniques generated robust cross-modal embeddings, significantly enhancing retrieval precision.
This work underscores the critical role of sophisticated fusion methods in aligning multimodal data, ensuring seamless retrieval and recommendation processes across text and image modalities.
Following the rise of Transformer~\cite{10.5555/3295222.3295349} as a prominent technique in various research fields, AlRahhal et al.~\cite{9925582,9840015} explored their application in image retrieval, proposing a multilanguage image-text model for remote sensing image retrieval. Their approach employs transformer-based encoders to extract modality features from image-text pairs and uses global average pooling to create a common similarity space. During training, the model optimizes contrastive classification losses for both text-to-image and image-to-text pairs, ensuring effective learning from both modalities. 
In a related study~\cite{9840015}, the authors addressed cross-modal retrieval challenges in remote sensing applications by using separate encoders for textual and visual features. Their system supports both Arabic and English captions in the text modality. They investigated two paradigms for language alignment: one where Arabic and English are learned independently, and another where both languages are learned jointly. Experiments on two cross-modal datasets showcased the promising capabilities of their approach.

\subsubsection{Speech Recognition} 
Since speech is the most effective and common form of human communication in daily life, academics have been concentrating on Automatic Speech Recognition (ASR) tasks over time, particularly in low-resource languages such as Arabic. 

Alrashoudi et al.~\cite{Alrashoudi2025} present a novel framework for mispronunciation detection and diagnosis (MDD) for non-native Arabic speakers. As the first stage of their framework, they use an ASR model by fine-tuning Wav2Vec2.0~\cite{9585401}, Hubert~\cite{9585401}, and Whisper~\cite{radford2022robustspeechrecognitionlargescale} to autoregressively predict phoneme-based transcriptions from the raw data, which includes audio and textual data.
They are creating their L2-KSU dataset, which includes 4086 utterances and six hours and six minutes of audio recordings associated with its transcriptions.

Another application where speech is essential is speech recognition for robot control.
Researchers enable robots to process speech as a natural medium for receiving human instructions. Sagheer~\cite{s_alaa2015}  proposed an Audio-Visual Speech Recognition (AVSR) system composed of three main components: face-mouth feature detection, automatic user identification, and a visual speech recognition module. The face-mouth feature detection uses the Viola-Jones detection module~\cite{990517}, which relies on integral image concepts to reduce computational costs via attentional cascade classifiers. For user authentication, a machine learning-based automatic identification system is implemented to address security concerns. 
The visual speech recognition module captures visemes from lip movements, with feature extraction and recognition carried out using an ANN-based Self-Organizing Map (SOM) for unsupervised competitive learning. For viseme feature recognition, k-NN and Hidden Markov Model classifiers are employed. These visual features are integrated with speech features extracted through Mel Frequency Cepstral Coefficients (MFCC)-based Automatic Speech Recognition (ASR).
Additionally, Sagheer introduced the Audio-Visual Arabic Speech (AVAS) database, containing $36$ isolated words and $13$ casual phrases collected from $50$ Arabic-speaking subjects, serving as a benchmark dataset for AVSR research.

\subsubsection{Video Classification} Dandashi et al.~\cite{7881438} proposed a system to classify news videos into categories such as shootings or explosions. The system starts by compiling an Arabic dataset and recognizing events, persons, dates, and locations from video speech transcriptions. For audio analysis, it employs classification tools like Yaafe and Essentia~\cite{10.1145/2502081.2502229}. These recognized events inform the visual module, directing it to focus on key video frames for more precise visual classification. The system employs effective fusion techniques, including feature-level, decision-level, and hybrid correlations, as well as weighted fusion methods, to enhance the contribution of each modality to the final classification decision.  
The authors, in a separate study~\cite{10.1007/978-3-319-89914-5_3}, combined both audio-based and textual-based methods to improve classification accuracy. The textual modality utilized key elements extracted through Named Entity Retrieval (NER) techniques, while the audio modality analyzed acoustic events such as clanking, scraping, and children's voices. The NER outputs and audio-based results were fused with visual-based classification outcomes using multimedia fusion classifiers to determine the final classification. The fusion strategy involved assigning set weights to each modality and using a preset threshold to ensure accurate classification. Furthermore, the study introduced a new Arabic dataset comprising Arabic news broadcast videos and raw news-related videos, establishing a baseline for future classification research. The work also leveraged a multilingual dataset to broaden its applicability.
While the research primarily relied on traditional rule-based methods for data representation and classification, the hybrid decision-level fusion approach effectively integrated modalities. However, adopting advanced data-driven techniques, such as deep learning-based multimodal fusion models, could further improve classification accuracy and adaptability to complex scenarios.

\subsubsection{Propaganda Detection} To address the spread of propaganda on the global internet,  particularly in the Arab world, Hasanain et al.~\cite{hasanain-etal-2024-araieval} organized the ArAIEval shared task as part of the ArabicNLP 2024 conference. The shared task comprises two components: (i) detecting propagandistic textual spans and identifying persuasion techniques in tweets and news articles, and (ii) distinguishing between propagandistic and non-propagandistic memes. This initiative provides a platform for evaluating methodologies in propagandistic content detection, focusing on both textual and multimodal data.
Notably, in the multimodal challenge, 11 teams submitted system description papers among which, we can cite the following:~\cite{haouhat-etal-2024-modos,zaytoon-etal-2024-alexunlp,wang-markov-2024-cltl-araieval,shah-etal-2024-mememind}. Most teams employed fine-tuning of transformer-based models, such as AraBERT, for handling textual inputs. These systems were complemented by various visual models, including CLIP, SAM, and ResNet, underscoring the growing importance of integrating textual and visual modalities for the effective detection of propagandistic content.
Haouhat et al.~\cite{haouhat-etal-2024-modos} submitted their work on propaganda detection in the ArAIEval shared task. They proposed a preprocessing pipeline for raw data before feeding it into the model. First, they utilized the Segment Anything Model (SAM) for image segmentation of meme images, followed by the application of the state-of-the-art image encoder CLIP to extract image embeddings. Finally, for multimodal classification, they leveraged Long Short-Term Memory (LSTM) networks to combine both textual and visual modalities, allowing the model to effectively handle the integration of different data sources.
This work emphasizes the importance of preprocessing in propagandistic meme detection. By using SAM and CLIP, they ensure a high-quality extraction of both visual features and image content. However, there is potential for enhancing multimodal fusion by exploring more advanced fusion techniques beyond LSTM, such as weighted attention-based methods, which could improve the interaction between the textual and visual components of the data.
AlexUNLP-MZ Team~\cite{zaytoon-etal-2024-alexunlp} participated in the ArAIEval task on propaganda detection by employing a Large Language Model (LLM) to extract features from the dataset. They considered both 'weighted loss' and 'contrastive loss' during training to optimize performance. For text classification, they utilized the BLOOMZ model~\cite{muennighoff-etal-2023-crosslingual}, a powerful LLM. For image classification, they experimented with CNN-based architectures, such as ResNet and DenseNet. In the multimodal setting, they combined the strengths of the BLOOMZ-1b1 model for text processing with ResNet101 for image classification to create a robust multimodal fusion system.
This approach reflects a comprehensive use of both language and image models, with the combination of BLOOMZ and ResNet101 providing a solid framework for tackling propaganda detection. However, the fusion method could be further enhanced by exploring more sophisticated attention mechanisms that could allow for more dynamic interactions between the text and image modalities, potentially improving classification accuracy.
CLTL Team~\cite{wang-markov-2024-cltl-araieval} applied a variety of models for text classification in the ArAIEval task on propaganda detection. Specifically, they utilized MARBERT, CAMeLBERT, and GigaBERT. The findings reveal that CAMeLBERT outperforms the others in terms of performance. For the visual modality, they experimented with EVA and CAFormer, two powerful models for image processing. In their multimodal approach, the team fused the embeddings of the text and image modalities using a multilayer perceptron (MLP) technique to create a unified representation for classification.
The team's use of CAMeLBERT for text classification demonstrates a thoughtful choice, as it is optimized for Arabic text, giving it an edge in handling linguistic nuances. The incorporation of EVA and CAFormer for image classification shows a solid attempt to leverage state-of-the-art image models. However, while the MLP fusion technique is effective, exploring more advanced fusion strategies like attention-based models could offer improved synergy between the text and image features, enhancing the overall performance.
MemeMind Team~\cite{shah-etal-2024-mememind} employed innovative data augmentation strategies to enhance their system's performance in the ArAIEval shared task on propaganda detection. For text data, they utilized synthetic text generated by GPT-4 (OpenAI, 2023) to expand the dataset. For the image modality, they leveraged DALL-E-2 to create augmented visual data and fine-tuned ResNet50, EfficientFormer (v2), and ConvNeXt-tiny architectures for image classification. In their multimodal approach, they fused the features of ConvNeXt-tiny and BERT to integrate textual and visual information effectively.
The use of GPT-4 and DALL-E-2 for data augmentation showcases the team’s innovative approach to tackling the challenge of limited labeled data in multimodal tasks. However, while the fusion of ConvNeXt-tiny and BERT provides a solid baseline, exploring dynamic fusion methods, such as cross-attention mechanisms, could further enhance the interaction between text and image modalities, potentially improving classification accuracy.

\subsubsection{Language Recognition} In the context of language identification, Alharbi~\cite{ALHARBI2024102084} introduced a multimodal framework combining audio and textual data for detecting Saudi dialects. The textual transcription is encoded using a BERT-based encoder, while the audio input is processed through a CNN-based acoustic model to extract relevant features. The authors utilize a self-attention mechanism to capture semantic relationships between the two modalities. Additionally, they design a neural network to align textual information with corresponding audio segments, ensuring effective representation and synchronization between modalities.
This work adopts an early fusion approach to combine audio and textual embeddings, effectively bridging the modalities. The alignment mechanism between textual and audio representations is particularly noteworthy, as it enhances the framework’s ability to handle the inherent temporal and semantic differences between the modalities.

\subsubsection{Speaker Role Recognition} Mehra et al.~\cite{Mehra2024-gc} introduced a novel strategy for spoken word recognition in Arabic, particularly under the constraints of a multilingual and low-resource setting. Their approach utilized the pre-trained Arabic Large xlsr-Wav2Vec2-53~\cite{grosman2021xlsr53-large-arabic} transformer model for speech-to-text conversion, processing text in both Buckwalter transliterations and Arabic script. It leveraged phonetic representation through the CMU dictionary\footnote{http://www.speech.cs.cmu.edu/cgi-bin/cmudict} and applied a grapheme-to-phoneme model for Buckwalter transliterations~\cite{BISANI2008434}. For Arabic script, stemming and unigram-based conversions are followed by FastText~\cite{joulin2016fasttext} word embeddings for vector representation~\cite{bojanowski2016enriching}. These vectors are processed by a three-layer dense model with batch normalization, and the final output is an average of results from both forms.

\subsubsection{Understanding and Generation} The rise of Large Language Models (LLMs) has recently garnered significant attention across various research domains, especially within multimodal systems. Although LLMs have achieved remarkable success in processing and understanding text, their potential to integrate and interpret multimodal data, such as text, audio, and images, has seen less attention, particularly within Arabic contexts. This is due to the complexities of the Arabic language, its dialects, and the scarcity of large annotated multimodal datasets. Despite these challenges, there have been a few notable efforts to apply Large Multimodal Models (LMM) to Arabic, specifically in areas such as sentiment analysis, text recognition, and language generation.
Bhatia et al.~\cite{bhatia-etal-2024-qalam} introduced Qalam, a novel LMM, designed specifically for Optical Character Recognition (OCR) and Handwriting Recognition (HWR) in Arabic script. This model integrates a vision transformer-based encoder with a transformer-based decoder, where the encoder captures visual information, and the decoder transcodes it into meaningful textual output. The model leverages autoregressive training, where it predicts the next token based on the previous ones, ensuring smooth text generation.
One of the key contributions of this work is the introduction of the Midad Benchmark, a dataset specifically crafted for Arabic OCR and HWR, allowing for a thorough evaluation of the model's performance. The authors compare Qalam with several well-established OCR models, including CRNN, Gated-CNN-BiLSTM-CTC, Tesseract, and TrOCR, demonstrating that Qalam outperforms these baselines in terms of accuracy and efficiency.

Alwajih et al.~\cite{alwajih-etal-2024-peacock} introduced Peacock, a cultural-aware Arabic LMM. The model is trained on a dataset created through the translation and filtering of English-based Multimodal datasets, ensuring the representation of various Arabic dialects. Peacock’s architecture follows the LLaVa $1.5$ model, using a CLIP encoder to process images, which are then embedded into the AraLLAMA $7$B input space through a two-layer MLP. Only the MLP layers are trained, while the rest of the model remains frozen. This approach offers efficiency but leaves room for future enhancements when more diverse datasets become available. Overall, Peacock provides a promising step towards improving Arabic LMM's with cultural sensitivity.

Alwajih et al.~\cite{alwajih-etal-2024-dallah} presented Dallah, a dialect-aware Arabic LMM. Similar to their previous work, the authors build a dataset by translating and filtering English-based datasets, using Google Translate and a threshold filtering method to ensure high-quality translations, while also covering a broad range of Arabic dialects. Dallah's architecture follows the LLaVa $1.5$ model, using a CLIP encoder to process images and project visual patch sequences into the AraLLAMA $7$B input space. Textual tokens are generated through the AraLLAMA tokenizer. The model trains only the two MLP layers that project the CLIP embeddings to the AraLLAMA model, leaving the rest of the structure frozen. This setup allows for efficient training but presents an opportunity for future improvements once more comprehensive datasets are available. Dallah is another significant step toward developing more accurate and culturally sensitive Arabic large multimodal models.
 
%%%%%%%%%%%%%%%%%%%
Fanar~\footnote{\url{https://fanar.qa/en}} is a recent  Arabic-Centric Multimodal Generative AI Platform. It supports language generation, speech-to-text recognition, and image generation tasks~\cite{fanarteam2025fanararabiccentricmultimodalgenerative}. The authors emphasize cultural preferences in the MENA region and they empower the model capabilities aligned with their objectives using many techniques, starting with high-quality data curation, designing the MorphBPE tokenization method~\cite{asgari2025morphbpemorphoawaretokenizerbridging}, maintaining the knowledge-preference coherence, and integrating the Retrieval Augmented Generation feature.

AIN~\cite{heakl2025ain}, the {Arabic Inclusive Multimodal Model} is designed to excel in diverse domains. AIN is a bilingual English-Arabic LMM designed to excel in both  English and Arabic, leveraging carefully constructed $3.6$ million high-quality Arabic-English multimodal data samples. It includes $8$ domains within $38$ sub-domains, including multi-image understanding, complex visual perception, handwritten document understanding, video understanding, medical imaging, plant diseases, and remote sensing-based land use understanding. AIN demonstrates strong performance with the $7$B model. It presents a significant step toward empowering Arabic speakers with advanced multimodal generative AI tools in various applications.

It is important to note that Arabic-based LMMs have benefited from benchmarking efforts aimed at understanding their performances. For instance, CAMEL-Bench~\cite{ghaboura2024camelbenchcomprehensivearabiclmm} is an open-sourced  multimodal framework that comprises around more than $29$K  questions that are
filtered from a larger pool of samples, where the quality is manually verified by native speakers to ensure reliable model assessment. 

\subsection{Approaches}
\label{sec:approaches}
The progression of multimodal learning techniques has transitioned from traditional probabilistic models to classical machine learning methods and, more recently, to the widespread adoption of deep learning architectures. 
Each methodological category has introduced distinct strengths and limitations, significantly influencing the design of multimodal datasets and their applications. This section examines the reviewed studies by organizing them into three categories: traditional, classical machine learning, and deep learning-based approaches. We highlight the foundational principles of each category and discuss specific studies that employ these techniques. 

\subsubsection{Traditional Methods}

Early studies relied on traditional methods because of their simplicity and effectiveness in structuring feature spaces. Rule-based approaches, in particular, use predefined logical rules or conditions--often derived from expert knowledge or heuristics-- to make decisions or classify data. To the best of our knowledge, we found only one study that used traditional methods in the Arabic context, which is demonstrated in~\cite{7944543}. Karkar et al.~employ a heuristic mapping process (i.e., basic matching techniques) to retrieve relevant images. Additionally, they utilized cosine similarity measures to improve the retrieval accuracy.

Hidden Markov Models (HMMs) emerged as a prominent approach for capturing sequential dependencies by modeling systems as a series of observable events governed by hidden states. Their probabilistic state transitions enabled an effective representation of temporal patterns. For example, in viseme-based speech recognition, HMMs were combined with k-Nearest Neighbors (k-NN) to improve classification accuracy by utilizing hidden state transitions alongside neighborhood-based voting techniques~\cite{s_alaa2015}.

Although traditional methods laid a solid groundwork for multimodal research, they were limited by their dependence on manual feature engineering, which required extensive domain expertise. Moreover, these methods often lacked scalability and robustness in dynamic environments, prompting the adoption of classical machine-learning approaches to overcome these challenges.

\subsubsection{Classical Machine Learning Methods}

Classical machine learning methods marked a significant step forward by integrating advanced algorithms capable of handling various tasks. These methods leveraged statistical and supervised learning techniques, enabling the integration of structured features from different modalities.

Support Vector Machines (SVMs) are a popularly used technique that employs supervised learning to find an optimal hyperplane that separates data points into distinct classes. By maximizing the margin between the support vectors, the SVMs provided robust classification capabilities. These models were often combined with other classical methods, such as Logistic Regression and L2 norm regularization, to enhance multimodal data processing, as demonstrated in various studies~\cite{ALROKEN2023103005,9677274,7881438}.

K-Nearest Neighbors (k-NN), a simple yet effective algorithm, classified data points based on the majority vote of their closest neighbors. When integrated with HMMs, as seen in viseme recognition applications, k-NN contributed to better classification accuracy in multimodal datasets~\cite{s_alaa2015}. 

Kernel methods, which map data into higher-dimensional spaces using kernel functions,  proved highly effective in addressing nonlinear problems by making them linearly separable. Techniques such as kernelized SVMs (KSVMs), which employed linear kernels for efficient binary classification, demonstrated improved performance in multimodal contexts~\cite{9148603}.

Lastly, Multilayer Perceptrons (MLPs) demonstrated their ability to capture nonlinear patterns in multimodal data. These models proved to be effective for tasks that require complex relationships between inputs from different modalities~\cite{10.1007/s10772-022-09981-w,bhatia-etal-2024-qalam}. However, MLPs faced limitations when dealing with sequential or spatial data, where models such as Recurrent Neural Networks (RNNs) or Convolutional Neural Networks (CNNs) were more adept at capturing complex dependencies.

While classical machine learning models expanded the scope of multimodal tasks, their reliance on feature engineering limited adaptability. This paved the way for deep learning architectures that automated feature extraction and supported end-to-end learning, marking a significant evolution in the field.

\subsubsection{Deep Learning Methods}

The advent of deep learning marked a transformative era for multimodal research. It enabled the development of sophisticated architectures that automated feature extraction and effectively captured complex interactions between modalities. This paradigm shift allowed researchers to handle unstructured and high-dimensional data more efficiently, leading to significant advancements in multimodal tasks.  

Convolutional Neural Networks (CNNs) became a cornerstone for processing visual data because of their ability to extract spatial features from images and video frames. CNNs were employed as versatile feature extractors and representation tools, playing an essential role in handling tasks such as object detection and image classification. Their adaptability made them a common choice for multimodal applications, as demonstrated in studies utilizing CNNs for cross-modal representations and feature extraction across diverse modalities~\cite{Zater2022BenchMarkingAI,8717419,ALROKEN2023103005}.  

Recurrent Neural Networks (RNNs) and their enhanced variant, Long Short-Term Memory Networks (LSTMs), addressed the challenges of sequential data, such as audio or text. LSTMs, in particular, excelled in capturing temporal dependencies by maintaining long-term context, which proved invaluable for tasks like image caption generation~\cite{Elbedwehy2023,cmc.2024.048104,Zater2022BenchMarkingAI}. Similarly, they were instrumental in speech and lip-reading tasks on the AVAS dataset, where their ability to handle variable-length sequences offered a robust framework for modeling temporal patterns.  

The emergence of Transformers revolutionized the way sequential data was processed. By leveraging attention-based mechanisms, transformers efficiently captured long-term dependencies and complex correlations between multimodal inputs. This attention-centric approach was further enhanced through the integration of cross-attention modules, enabling nuanced interaction across modalities. Transformers have been widely adopted for fine-grained cross-modal understanding~\cite{zaytoon-etal-2024-alexunlp,ALHARBI2024102084}.  

Building on the foundation of transformers, Large Language Models (LLMs) emerged as cutting-edge tools for multimodal research. Some studies employed LLMs as synthetic data generators, enhancing model performance through diverse and high-quality samples~\cite{shah-etal-2024-mememind,alwajih-etal-2024-peacock}. Additionally, autoregressive training paradigms enabled smoother and more coherent outputs\cite{10.5555/3295222.3295349,Radford2018ImprovingLU}, especially in text generation tasks~\cite{bhatia-etal-2024-qalam}. Unique applications of transformers in Arabic contexts included efficient speech recognition, even under the constraints of multilingual and low-resource settings~\cite{Mehra2024-gc}.  

Deep learning methods have revolutionized the multimodal research landscape, offering tools capable of learning complex representations and capturing intricate relationships across diverse data types. Despite these advancements, challenges persist, including significant computational demands, the scarcity of data in low-resource settings, and the ongoing need to enhance model interpretability. These areas remain at the forefront of active research.

\subsubsection{Hybrid Methods}
 
Hybrid models take advantage of the strengths of multiple approaches, combining deep learning architectures with traditional methods to enhance multimodal analysis. For instance, some studies integrate CNNs for spatial feature extraction from images with LSTMs to process temporal dependencies in sequential data\cite{6795963}. Others employ Transformer layers to enable cross-modal interactions, such as combining advanced models like SAM and CLIP with LSTMs to tackle diverse multimodal tasks effectively. These combinations showcase the versatility of hybrid models in addressing the challenges posed by different data modalities~\cite{haouhat-etal-2024-modos,alwajih-etal-2024-peacock}.

A notable approach involves pairing deep learning feature extractors, such as Transformers, with traditional statistical methods for multimodal classification. For example, similarity matrices constructed from image-text contrastive learning have been employed to enhance classification accuracy~\cite{9840015, 9925582, 10453875}. This combination allows for capturing nuanced relationships between modalities while benefiting from the structured decision-making process inherent in classical statistical methods.

Other hybrid techniques include back-translation for augmenting multimodal datasets~\cite{alwajih-etal-2024-peacock}. Furthermore, some studies use frozen pre-trained models for feature extraction, subsequently integrating classical neural layers to adapt these features to task-specific requirements~\cite{alwajih-etal-2024-dallah}. Such strategies enable the reuse of large-scale pre-trained models, reducing computational costs while achieving high accuracy on specific tasks.

Hybrid methods have proven particularly effective in scenarios with limited resources or where single-model architectures fall short. By integrating diverse methodologies, these approaches push the boundaries of multimodal learning, offering flexible solutions to a range of tasks.

\subsection{Challenges}
\label{sec:challenges}
Multimodal Machine Learning has made remarkable strides and demonstrated broad applicability. Nevertheless, substantial challenges remain, particularly in Arabic language applications. Key aspects requiring further enhancement include fusion methods, modality translation, alignment strategies, and data representation. The reviewed papers have made progress in addressing some of these issues.

\subsubsection{Fusion Methods}

Multimodal learning tasks rely on the effective combination of modalities, such as visual, audio, and textual embeddings, to create a unified representation that drives system performance. In the literature, fusion techniques are classified into three main types: early, late, and hybrid. 

Early fusion unifies the representations of multiple modalities at an initial stage, typically before the prediction phase. For example, features are often combined and fed into dense neural networks with batch normalization layers to ensure stability and efficient training~\cite{Mehra2024-gc}. This method enables the model to capture interactions between modalities early in the pipeline, potentially improving the coherence of multimodal representations.

Early fusion employs techniques such as simple vector operations (e.g., concatenation, addition, or weighted vectors~\cite{7881438}). Furthermore, more sophisticated methods such as cross-modal fusion, which is based on mechanisms of cross-attention, are available to capture semantic relationships between modalities, facilitating dynamic and adaptive representations~\cite{bhatia-etal-2024-qalam}.

In contrast, late fusion combines predictions or high-level features from each modality after individual processing. Decision fusion is a prominent example of this approach, where the outcomes of unimodal predictors are integrated at the decision-making stage to improve performance~\cite{10.1007/s10772-022-09981-w}. Late fusion is advantageous in scenarios where each modality's individual contribution needs to be preserved and analyzed independently.

Hybrid fusion blends the benefits of both early and late fusion methods within a unified framework. For example, Group Gated Fusion (GGF) dynamically integrates information from different modalities~\cite{liu2022groupgatedfusionattentionbased}, ensuring effective interaction while leveraging the complementary strengths of early and late fusion~\cite{10453875}. Similarly, hybrid multi-level fusion approaches capture cross-modal dependencies at various levels, enabling nuanced and context-aware representations~\cite{9148603}.

\begin{figure}[h!]
    \centering
       \includegraphics[width=\textwidth]{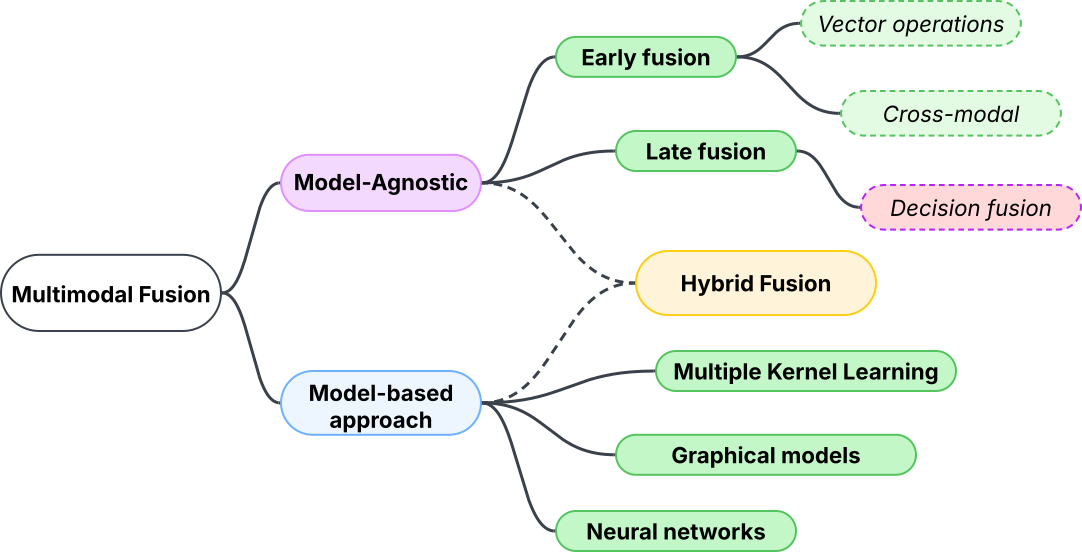}
    \caption{An Overview of Fusion Mechanisms in Multimodal Machine Learning.}
    \label{fig:taxonomy-fusion} 
\end{figure}

%noisy and missing data 
Some works classified multimodality fusion approaches into more detailed classification, as illustrated by Figure~\ref{fig:taxonomy-fusion}. The latter highlighted two main categories: model-agnostic and model-based approaches~\cite{10.1109/TPAMI.2018.2798607}.
Model-agnostic approaches do not depend on a specific machine learning model. Instead, they leverage general techniques to integrate multimodal data as we described earlier.

Model-based approaches explicitly incorporate multimodal fusion mechanisms into their architecture. These primary include, first, Kernel-based methods, extend support vector machines (SVMs) by utilizing different kernels for each modality, enabling better fusion of heterogeneous data (e.g., such as Multiple Kernel Learning (MKL)). Secondly, graphical models such as Bayesian networks, which are based on probabilistic frameworks, are especially well-suited for temporal and sequential multimodal tasks. Third, Neural Networks-based methods, which leverage modern deep neural architectures, have proven highly effective for multimodal fusion.

In general, all these fusion categories are designed to integrate information from all available modalities, even in the presence of missing or noisy inputs. Researchers have addressed these challenges through various approaches, such as imputing missing data using generative models, designing noise-resilient architectures, or employing robust late fusion techniques that weigh modality contributions based on reliability. Furthermore, exploring self-supervised and contrastive learning techniques to improve cross-modal alignment and representation learning significantly enhances fusion robustness in case of incomplete or noisy inputs.

\subsubsection{Modality Translation}

Modality translation refers to the process of switching or generating one modality from another. This technique plays a crucial role in bridging gaps between modalities, making it possible to derive enriched insights and enabling a broader spectrum of applications. The process can generally be classified into two categories: unimodal and multimodal settings.

In unimodal settings, the model operates on a specific input modality and transforms it into a different modality. For example, textual inputs can be processed through an LSTM decoder to produce corresponding output tokens, as demonstrated in tasks like image captioning~\cite{Elbedwehy2023, cmc.2024.048104}. This category is often leveraged in scenarios where the transformation involves single-modality data, such as converting visual input into another structured textual text.

In contrast, multimodal settings embrace unified designs capable of representing and translating between multiple modalities. These approaches rely on advanced architectures that harmonize diverse modalities, such as textual and visual inputs. Models like CLIP~\cite{radford2021learningtransferablevisualmodels}, LLAMA-3~\cite{grattafiori2024llama3herdmodels}, and ChatGPT-4~\cite{openai2024gpt4technicalreport} exemplify this paradigm. For instance, CLIP's visual output matrix has been utilized to match meme images with their corresponding textual descriptions, showcasing the effectiveness of multimodal translation in capturing nuanced relationships~\cite{haouhat-etal-2024-modos}. This approach improves tasks such as image captioning, video analysis, and multimodal sentiment analysis by creating seamless interaction across modalities.

Such advancements in modality translation empower complex tasks like visual question answering, multimodal dialogue systems, and creative content generation, where the interplay of modalities enhances overall system performance.

Tadas et al.~\cite{10.1109/TPAMI.2018.2798607} classify multimodal translation techniques into two approaches:
\textbf{Example-based approaches}: These rely on similarity functions to identify the closest instance to the input modality within a source space. In essence, this is a retrieval-based system where paired input and target translated samples are used. 

\textbf{Generative-based approaches}: These construct the target modality from scratch, often utilizing encoder-decoder architectures. This method extends language generation techniques to handle not just textual modalities but also multimodal translation tasks.  For example, an encoder-decoder model might generate descriptive captions for images by encoding visual features and decoding them into coherent text.
Despite significant progress, multimodal machine translation faces challenges, particularly subjectivity in translation, where multiple correct outputs may exist for tasks like image captioning, visual question answering, and others. Addressing this subjectivity remains a critical area for further research.

\begin{figure}
\centering
\begin{subfigure}{0.45\textwidth} 
  \centering
  \includegraphics[width=\textwidth]{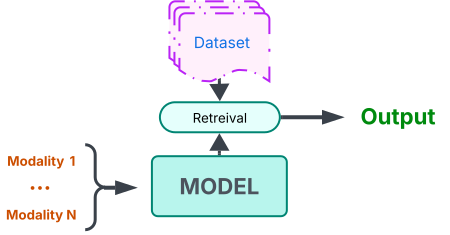}
  \caption{Generative-based Approach}
  \label{fig:sub1}
\end{subfigure}
\hfill
\begin{subfigure}{0.45\textwidth} 
  \centering
  \includegraphics[width=\textwidth]{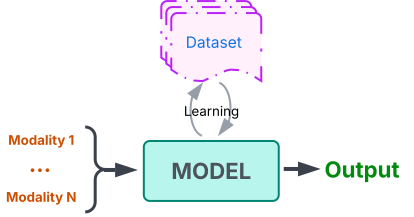}
  \caption{Example-based Approach}
  \label{fig:sub2}
\end{subfigure}

\caption{Overview of Multimodal Translation: This figure illustrates both approaches, where the input can consist of one or multiple modalities. For instance, a textual prompt can be fed into a Text-to-Image model to generate an image.}
\label{fig:trans}
\end{figure}

Figure~\ref{fig:trans} provides an overview of the multimodal translation process, illustrating the different approaches and categories into a unified translation framework.

\subsubsection{Modality Alignment}
Modality Alignment refers to the process of identifying relationships and establishing correspondences between various parts of all involved modalities. For example, in a tutorial video where a person discusses an educational topic, alignment involves matching the video frames with their corresponding textual descriptions of the subject concepts. Effective alignment is essential for ensuring semantic and temporal coherence across modalities, a cornerstone for the success of multimodal applications.

Researchers have employed diverse mechanisms to achieve alignment, including Neural Network Alignment, Cross-modal Mechanisms, and Graph Neural Network Alignment~\cite{muttenthaler2025humanalignmentneuralnetwork,kim2022crossmodalalignmentlearningvisionlanguage,SONG2024110484}.

Neural Network Alignment focuses on synchronizing textual information with corresponding audio segments. This approach addresses temporal mismatches and ensures synchronized multimodal representations. By integrating neural networks for alignment tasks, frameworks can better manage semantic and temporal differences, among the most significant challenges in multimodal tasks~\cite{ALHARBI2024102084}.

Cross-attention Mechanisms play a pivotal role in aligning visual and textual modalities. These mechanisms enable dynamic feature interactions, improving the model's ability to identify nuanced relationships between text and image representations. For example, cross-attention modules enhance accuracy by ensuring better alignment and integration of modalities, making them especially effective in tasks such as image captioning, video-text matching, and multimodal sentiment analysis~\cite{bhatia-etal-2024-qalam}.

Through these methods, researchers have significantly improved the alignment between modalities, enabling more accurate and context-aware multimodal systems.

\subsubsection{Data Representation}
Data representation is one of the main challenges in machine learning, as the way input data is represented significantly influences the system's performance. Data representation methods can be broadly categorized into two main approaches: handcrafted representations and learned representations.

Handcrafted representations involve traditional and statistical methods to extract relationships between the components of the modality. For textual data, techniques such as Bag of Words (BoW), Term Frequency-Inverse Document Frequency (TF-IDF), and N-grams are commonly used. For visual data, Histogram of Oriented Gradients (HOG), Scale-Invariant Feature Transform (SIFT)~\cite{Lowe2004}, and edge detection features are widely applied. Similarly, for audio and time series data, signal processing-based techniques like Mel-Frequency Cepstral Coefficients (MFCCs), spectrograms, and Fourier transform features are effective. For instance,~\cite{7944543} employs an entity-word matrix composition approach to identify semantic connections using statistical methods. For acoustic modalities, Authors of~\cite{7881438,10.1007/978-3-319-89914-5_3} implement MFCC and spectral features to analyze audio data, while other studies explore facial expressions and movements to extract meaningful visual features~\cite{Mukeshimana23,khandait2012automaticfacialfeatureextraction}.

In contrast, learned representations leverage advanced machine learning techniques to provide more versatile and adaptive representations. These representations are derived by training unimodal or multimodal models on large datasets, ensuring context awareness and capturing complex, non-linear relationships in the data. This data-driven approach widely uses deep learning techniques such as deep neural networks, convolutional neural networks, recurrent neural networks, and transformers.  This approach also enables transfer learning, reusing pre-trained models for related tasks. Researchers frequently use pre-trained models such as AraBERT, MARBERT, ResNet, and VGG to effectively represent data inputs~\cite{Elbedwehy2023, cmc.2024.048104, 7944543,Zater2022BenchMarkingAI}.

Table~\ref{tab:data_representation} highlights data representation methods, providing references pertinent to techniques used in Arabic multimodal machine learning.
\begin{table}[H]
\centering
\caption{Overview of data representations in multimodal machine learning. This table categorizes various modalities—Textual, Visual, Audio, and Multimodal—into handcrafted and learned representations. It highlights key methods and pre-trained models for each modality, providing the references for understanding the techniques used in Arabic MML. }
\label{tab:data_representation}
\scriptsize
\begin{tabular}{C{1 cm} p{ \dimexpr\textwidth - 4cm\relax } C{1.5cm} } 
\toprule
\textbf{Mod} & \textbf{   Representation} & \textbf{Ref} \\ \midrule

\multirow{4}{*}{\rotatebox{90}{     Textual} }
& \textbf{Handcrafted:} Bag of Words (BoW), Term Frequency-Inverse Document Frequency (TF-IDF), N-grams. & \cite{7944543} \\  \\ 
& \textbf{Learned:} Pretrained models such as AraBERT, MARBERT; Transfer learning from language models. & \cite{Zater2022BenchMarkingAI,cmc.2024.048104} \\ \\ \midrule

\multirow{5}{*}{\rotatebox{90}{      Visual} }
& \textbf{Handcrafted:} Histogram of Oriented Gradients (HOG), Scale-Invariant Feature Transform (SIFT), Edge detection features. & \cite{7881438} \\ \\ 
& \textbf{Learned:} Pretrained models such as ResNet, VGG; Fine-tuned CNNs for specific tasks. & \cite{cmc.2024.048104,Elbedwehy2023} \\ \\  \midrule

\multirow{4}{*}{\rotatebox{90}{ Audio} }
& \textbf{Handcrafted:} Mel-Frequency Cepstral Coefficients (MFCCs), Spectrograms, Fourier transform features. & \cite{7881438,10.1007/978-3-319-89914-5_3} \\ \\ 
& \textbf{Learned:} Learned spectral embeddings using RNNs or Transformers; Transfer learning from audio models. & \cite{cmc.2024.048104} \\\\  \midrule

\multirow{4}{*}{\rotatebox{90}{ Multimodal} }
& \textbf{Handcrafted:} Manual feature engineering to align modalities. & -- \\  \\ 
& \textbf{Learned:} Unified multimodal embeddings, Cross-modal representations (e.g., CLIP, SAM). & \cite{haouhat-etal-2024-modos} \\ \\  \bottomrule

\end{tabular}
\end{table}

\section{Discussion}
\label{sec:Discussion}
\begin{figure}
    \centering
\includegraphics[width=\textwidth]{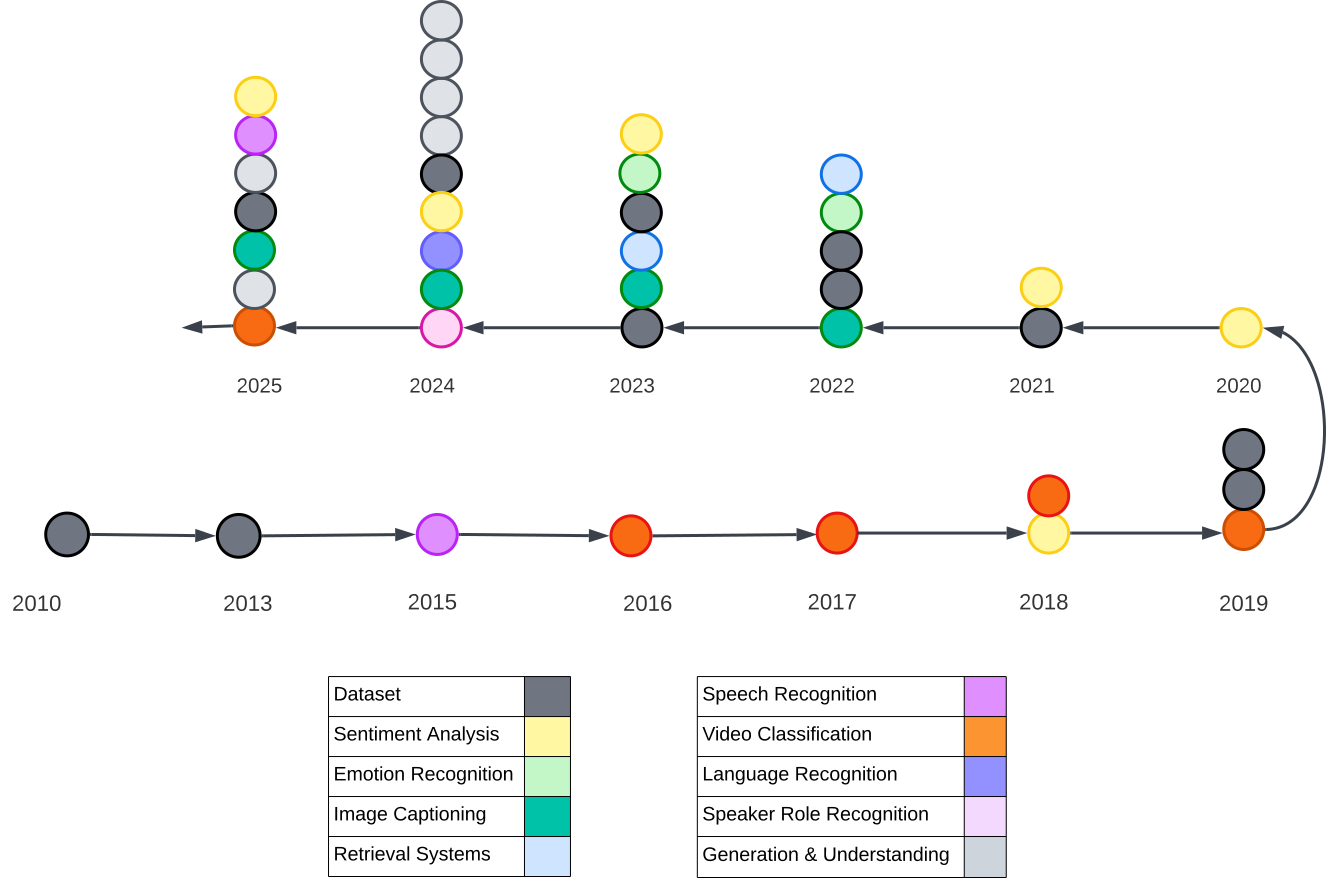}
\caption{Chronological Overview of the Surveyed Papers}
\label{fig:chrono_papers}
\end{figure}

In Figure~\ref{fig:chrono_papers}, we illustrate chronologically the  Arabic MML surveyed work to highlight the Arabic MML's progress over time.

% Chronological Taxonomy Table
%%%%%%%%%%%%%%%%%%%%%%%%%%

The first insights from this overview  reveal many:

\subsection*{Datasets are Prior}
%\textbf{Datasets are Prior. }
Around a quarter of the contributions focused on designing multimodal datasets, highlighting the scarcity of multimodal data available for Arabic AI-based multimodal systems. In fact, among the contributions, $12$ out of \NbSurveyed are dedicated to this area, emphasizing its significance in the field.
  
These efforts have also been complemented by developing various datasets to support multimodal research in the Arabic language. These studies are critical, given the lack of comprehensive resources available in comparison to other widely studied languages, such as English.

%%%%%%%%%%%%%%%%%%%%%%%%%%%%%%%%%%
\subsection*{Popular Targeted applications}
%\textbf{Popular Targeted Tasks}

We observed a diverse range of tasks that have been explored. However, sentiment analysis,  retrieval systems, and computer vision --video classification and image captioning-- are the most commonly targeted tasks. This focus is likely due to their broad applicability in real-world applications. 

\subsection*{Large Specter of Approaches}
 %\textbf{Large Specter of Approaches}
 
Researchers have transitioned from traditional methods to classical machine learning (ML) and, subsequently, to deep learning-based approaches, significantly enhancing the ability to model multimodal interactions. Traditional and classical models, such as Support Vector Machines (SVMs), provided the foundation for basic modality integration. In contrast, modern deep learning methods, such as Transformers, enable more comprehensive multimodal learning, facilitating breakthroughs in complex tasks such as sentiment analysis, emotion recognition, and sign language processing.

\subsection*{MML Challenges less Addressed}
While the initiatives are valuable for establishing a baseline~\cite{asgari2025morphbpemorphoawaretokenizerbridging, haouhat-etal-2024-modos, Mehra2024-gc,9677274}, they fall short in addressing more complex and deep challenges that are central to advancing MML. In fact, despite these efforts, current research in Arabic MML remains largely intuitive and concentrated on relatively straightforward problems.  For instance,  \textit{Alignment Strategies}:  aligning modalities such as text, audio, and visuals, especially in the context of semantic and temporal synchronization, is an area where there is a clear lack of robust solutions.
\textit{Fusion Techniques}:
The critical task of combining multimodal inputs effectively, whether through early, late, or hybrid fusion, remains underexplored in the Arabic domain. Particularly in cases where data is missing or noisy in a real-world environment.
\textit{Efficient Architectures}: Developing optimized multimodal models, especially tailored for Arabic-specific large language models and transformers, remains a significant challenge that demands further exploration.
Addressing these challenges could lead to more scalable and efficient solutions.

These gaps underscore the need for Arabic MML research to evolve beyond dataset curation and application-focused studies to tackle fundamental technical challenges. Advancements in this area would not only enhance performance on existing tasks but also enable researchers to address more complex real-world problems where multimodal data integration is essential.

\subsection*{Reuse of Existing Models}

Most existing studies primarily rely on pretrained models—originally designed for other languages or general-purpose tasks—such as BERT, ResNet, or CLIP to represent Arabic content. These models are typically fine-tuned or adapted with minimal modifications for Arabic-specific contexts. Subsequently, simple combination methods, such as concatenation fusion or basic alignment mechanisms, are employed to integrate information from multiple modalities and perform downstream tasks.

In contrast, the Fanar team~\footnote{\url{https://fanar.qa/en}} adopted a comprehensive approach to developing Fanar Star from scratch, starting with data collection and progressing through the training stage. This method ensures a more robust learning process and deeper knowledge acquisition.

\subsection*{LMMs Emergence and Deployment}

Research on LMMs has progressed more slowly for Arabic compared to extensively studied languages like English. Notably, the first Arabic LMMs were introduced only in $2024$. This delay can be attributed to the relatively late development and maturation of Arabic unimodal LLMs, which began emerging around $2017$, focusing on both Modern Standard Arabic (MSA) and various dialects.

Some studies have employed unimodal LLMs as synthetic data generators to enhance performance in data-scarce settings. Others have utilized autoregressive training to predict the next token based on previous tokens, enabling smooth and coherent text generation. Additionally, LLMs have been leveraged for multimodal tasks, such as generating image captions or performing speech-to-text conversion in multilingual and low-resource environments. These capabilities highlight the versatility of LLMs in seamlessly bridging modalities with minimal supervision, further advancing MML.

\section{Conclusion} \label{sec:conclusion}
% proposed conclusion
The field of Arabic Multimodal Machine Learning (MML) has reached a certain level of maturity, making this survey both timely and impactful. By introducing a novel taxonomy centered on four key axes: data-centric, applications, approaches, and challenges, we have outlined a comprehensive landscape of Arabic MML research. This framework, grounded in Arabic Language Technology, provides a systematic approach to analyzing and advancing work in this domain. Our comprehensive overview of this field serves as a foundational guide for readers to better understand this challenging and rapidly evolving field.

Our analysis reveals a strong emphasis on the construction of multimodal datasets, with sentiment analysis, retrieval systems, and computer vision emerging as dominant applications. The transition from traditional methods to Machine Learning and Deep Learning techniques demonstrates the field's growing sophistication. However, most existing studies continue to rely on pre-trained models to represent Arabic content. Furthermore, despite recent advancements, core difficulties --such as alignment strategies, fusion methods, and the integration of complex multimodal interactions--remain underdeveloped. Addressing these gaps will require more inventive and resilient solutions to push the boundaries of Arabic MML research.

From a scientific perspective, this work highlights the critical importance of addressing the core challenges in MML, notably in attaining effective alignment and fusion between modalities. The proposed taxonomy provides a clear roadmap, helping researchers identify gaps and prioritize efforts--especially in leveraging LMMs and advancing fusion techniques. 

On a practical level, the insights from this survey can inform the development of more efficient and scalable multimodal systems, with substantial potential for applications in healthcare, education, and human-computer interaction, where Arabic multimodal data is becoming increasingly relevant.

Looking ahead, the field must focus on overcoming the persistent challenges of alignment and fusion while exploring the revolutionary potential of LMMs for Arabic-specific tasks. Interdisciplinary collaboration will be essential to address real-world problems and drive innovation. By building on existing and tackling outstanding concerns, Arabic MML can continue to flourish, contributing not only to academic achievements but also to practical solutions that benefit the Arabic-speaking world and beyond.

\bibliographystyle{plain}
\bibliography{ref}

\end{document}